\def\year{2022}\relax
\documentclass[letterpaper]{article} 
\usepackage{aaai22}  
\usepackage{times}  
\usepackage{helvet}  
\usepackage{courier}  
\usepackage[hyphens]{url}  
\usepackage{graphicx} 
\graphicspath{ {./Images/} }
\usepackage{subcaption}
\usepackage{natbib}  
\usepackage{caption} 
\DeclareCaptionStyle{ruled}{labelfont=normalfont,labelsep=colon,strut=off} 
\usepackage{algorithm}
\usepackage{algorithmic}

\usepackage{newfloat}
\usepackage{listings}
\floatstyle{ruled}
\newfloat{listing}{tb}{lst}{}
\floatname{listing}{Listing}
 \usepackage{hyperref} 

\title{On the Transferability of Pre-trained Language Models: \\
A Study from Artificial Datasets}
\author {
    Cheng-Han Chiang,
    Hung-yi Lee 
}
\affiliations {
    
     National Taiwan University,Taiwan
    \\
    dcml0714@gmail.com, hungyilee@ntu.edu.tw
}

\begin{document}

\maketitle

\begin{abstract}
Pre-training language models (LMs) on large-scale unlabeled text data makes the model much easier to achieve exceptional downstream performance than their counterparts directly trained on the downstream tasks. 
In this work, we study what specific traits in the pre-training data, other than the semantics, make a pre-trained LM superior to their counterparts trained from scratch on downstream tasks.
We propose to use artificially constructed datasets as the pre-training data to exclude the effect of semantics, and further control what characteristics the pre-training corpora have.
By fine-tuning the pre-trained models on GLUE benchmark, we can learn how beneficial it is to transfer the knowledge from the model trained on the dataset possessing that specific trait.
We define and discuss three different characteristics in the artificial dataset: 1) matching the token's uni-gram or bi-gram distribution between pre-training and downstream fine-tuning, 2) the presence of the \textit{explicit dependencies} among the tokens in a sequence, 3) the length of the \textit{implicit dependencies} among the tokens in a sequence. 
Our experiments show that the explicit dependencies in the sequences of the pre-training data are critical to the downstream performance.
Our results also reveal that models achieve better downstream performance when pre-trained on a dataset with a longer range of implicit dependencies.
Based on our analysis, we find that models pre-trained with artificial datasets are prone to learn spurious correlation in downstream tasks.
Our work reveals that even if the LMs are not pre-trained on natural language, they still gain transferability on certain human language downstream tasks once the LMs learn to model the token dependencies in the sequences. 
This result helps us understand the exceptional transferability of pre-trained LMs.  
\end{abstract}

\begin{figure*}[t]

\centering
\includegraphics[width=1.0\linewidth]{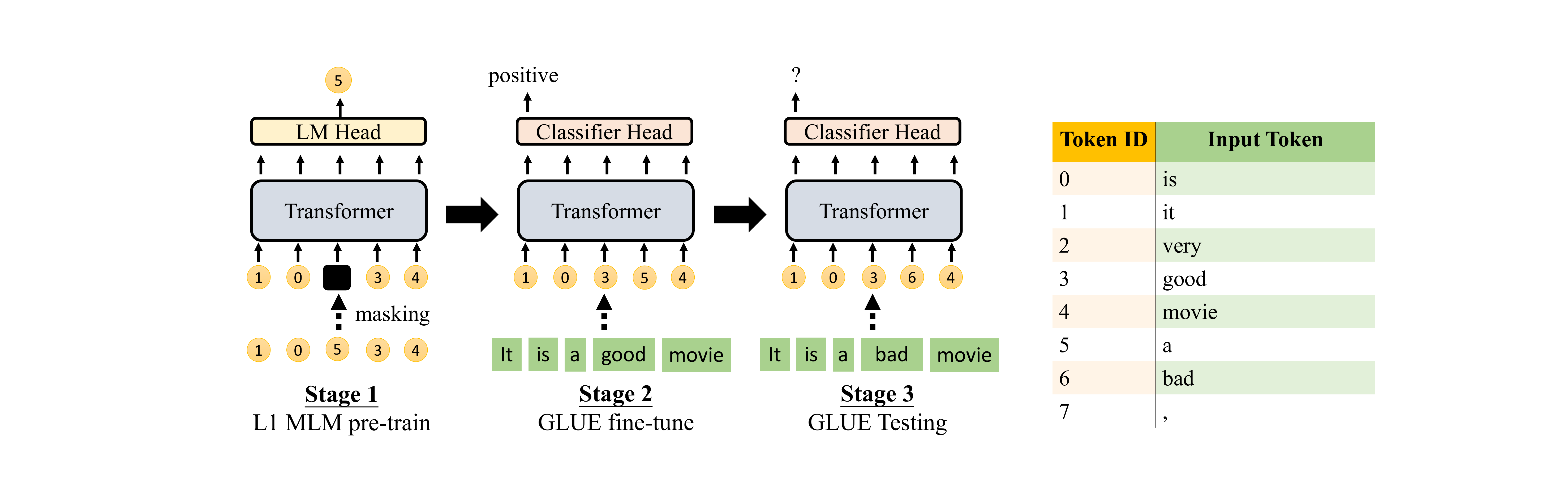}

\caption{ 
The workflow in our studies. 
\textbf{Stage 1}: Pre-training the whole MLM on language 1 (L1), in which each sentence is a sequence of token ID generated by certain rules.
\textbf{Stage 2}: Fine-tuning the whole model on English downstream tasks (GLUE tasks).
\textbf{Stage 3}: Evaluating the performance of the fine-tuned models on the English downstream tasks.
In stages 2 and 3, the model takes English token sequences as input.
To do that, each token embedding in the token embedding layer needs to be mapped to a token in English.
For example, we map the first token embedding in the embedding table, whose token ID is 0, to the English token "is"; and the second token embedding in the embedding layer, whose token ID is 1, to the English token "it". 
The whole process, from stage 1 to stage 3, takes three days on a single V100 GPU.
}
\label{fig:exp}
\end{figure*}

\section{Introduction}
Pre-training LMs by masked language modeling (MLM) is prevalent in the natural language processing (NLP) community, and they are indispensable to a variety of NLP tasks.
The popularity of pre-trained LMs mainly lies in their exceptional transferability on downstream tasks: fine-tuning these downstream-agnostic pre-trained models on miscellaneous downstream tasks often gives extraordinary performances compared with training from scratch.
While the exact reasons for the success of MLM is unclear, some have attributed this success to the pre-trained models having learned the semantic dependencies among tokens~\citep{saunshi2020mathematical} and being able to model the complex co-occurrence statistics of the tokens~\citep{sinha2021masked}.
While these justifications are reasonable, it is unclear whether there are other factors of the pre-training data that affect the transferability of the pre-trained LMs. 

The core problem we determine to answer is: What specific traits in the pre-training data, other than the semantics, make a pre-trained LM more easier to achieve better downstream performance.
To answer the above question, we design the following experiments: we create miscellaneous artificial datasets, each possessing different traits, and we pre-train many different transformer LMs on those datasets. 
We then fine-tune the pre-trained models on English downstream tasks.
The process is illustrated in Figure~\ref{fig:exp}. 
Illustrations of the artificial datasets used in this paper are in Figure~\ref{fig:data}.
This is the first paper to study a transformer-based LM's transferability through the lens of artificial datasets. 

Based on the experiments, we have the following takeaway observations: 
\begin{itemize}
\item We find that, surprisingly, pre-training MLMs on certain artificial datasets with no natural language semantics information makes their downstream task performance superior to models pre-trained from scratch on downstream tasks.
\item We discover that pre-training on data with a longer range of both\textit{ explicit} and \textit{implicit token dependencies}\footnote{The definition of explicit token dependency and implicit token dependency will be given in Section~\ref{sec:explicit} and Section~\ref{sec:long-term}, respectively.} makes them superior to their counterparts pre-trained on data with shorter token dependencies. This indicates that the ability to model long-term dependency among tokens in the sequence is important for the transferability of the pre-trained model.
\item We analyze the models' behaviors against two challenging datasets and show that models pre-trained with artificial datasets are vulnerable toward spurious correlation.
\end{itemize}

Supplementary materials are available at \url{https://github.com/d223302/Transformer-Structure}.

\section{Related Work} 
Our work uses artificial datasets to understand pre-trained LMs' ability to be fine-tuned.
~\citet{bhattamishra-etal-2020-ability} also use artificial dataset (formal language) to study a transformer model's behavior, but their work only involves the transformer model's ability to recognize certain types of formal languages.

Our work may seem to resemble the Test for Inductive Bias via Language Model Transfer (TILT)~\citep{papadimitriou-jurafsky-2020-learning} at first sight, which trains a long short-term memory (LSTM) LM on one language, which may be non-natural language, followed by only fine-tuning word embeddings on Spanish and test the perplexity on Spanish. 
In fact, this work is very different from TILT. 
The main purpose of TILT is to analyze the encoding of grammatical structure in LSTM LMs, so they do not fine-tune the LSTM on Spanish.
The setting of TILT does not match the common setting widely applied nowadays, in which we fine-tune pre-trained LMs on downstream tasks.
Different from TILT, our goal is to understand what trait of the pre-training datasets makes the fine-tuned model perform better than models trained from scratch on downstream tasks, thus we fine-tune the transformer LMs on the downstream tasks.  

~\citet{sinha2021masked} construct datasets from natural language corpora for pre-training by breaking the word order in a sentence.
Our work can be seen as a complementary work to theirs: ~\citet{sinha2021masked} break the word order while preserving the 'local statistics' of a sentence, thus preserving semantic features in the pre-training data to some extent. 
From their experiments, they show that purely distributional information (local co-occurrences of words) largely explains the success of MLMs.
Contrarily, we discard the semantics in the pre-training data while equipping the model with the ability to model token dependencies during pre-training.
In our experiments, we illustrate that part of the success of MLMs can be attributed to their ability to model token dependencies within a sequence.

\section{Analyzing LM's Transferability by Artificial Data}
\label{sec:setup}

The core idea of our experiment is to use different artificial datasets that have different traits to understand how a specific trait in the pre-training data affects the downstream performance.
In our experiments, we pre-train \emph{n} RoBERTa~\citep{liu2019roberta} models on \emph{n} different types of pre-training data.
Due to our constrained resources, we use RoBERTa-medium in our experiments; we will refer to RoBERTa-medium as RoBERTa in our paper.
We call the pre-training data L1 (first language).
The number of tokens in the pre-training corpora of all our L1s is around 100 million.
While the dataset is small, ~\citet{micheli2020importance} have shown that this amount of data is sufficient to pre-train a compact language model.

We then evaluate the pre-trained models' ability by fine-tuning them on different downstream tasks.
We adopt the GLUE ~\citep{wang2019glue, socher-etal-2013-recursive, dolan-brockett-2005-automatically, cer-etal-2017-semeval, williams-etal-2018-broad, rajpurkar2016squad} benchmarks to evaluate the models pre-trained on different L1s.
We exclude WNLI, RTE, and CoLA for their notoriously unstable performance~\citep{devlin2019bert,dodge2020fine}.
We use a specific set of hyperparameters and three different random seeds to fine-tune the model for each task. We report the average and standard deviation over different seeds of the results on the evaluation set. 
The overall workflow is illustrated in Figure~\ref{fig:exp}.
Details regarding all experiments can be found in Appendix~\ref{app: pretrain}.

\subsection{Artificial Datasets}
We construct different artificial datasets as L1 in this paper.
Each artificial dataset is constructed based on certain rules, which can be deemed as the grammar governing the artificial dataset.
Detailed construction and illustration of artificial datasets can be found in Section~\ref{sec:distribution} to ~\ref{sec:long-term} and Figure~\ref{fig:data}.
We design different grammar to scrutinize how the downstream performance may vary due to the characteristic of the pre-training data.
The vocabulary\footnote{Vocabulary is the set of tokens used by the language model, including special tokens such as \([MASK]\) and \([CLS]\).} of artificial datasets contains integers ranging from 0 to 29994 and 5 special tokens.
We choose to use 29995 tokens since the vocabulary size for our downstream English tasks is 29995 plus 5 special tokens.
We will use the word 'token', 'token ID', and 'integer' interchangeably to refer to the token for the artificial datasets. 
The sequence length of each sequence in the artificial pre-training data is determined by uniformly sampling between 100 and 120 (128 is the max sequence length the model we trained can process, and this is
sufficient for GLUE downstream tasks in most of the cases).

\subsection{Vocabulary during Pre-training and Fine-tuning}
The vocabulary for fine-tuning the English downstream tasks is obtained by Byte Pair Encoding (BPE), the vocabulary size is 30K, including five special tokens.
Since the vocabulary used during pre-training has no overlap with that during fine-tuning, the model cannot transfer any semantic knowledge from the pre-trained models.
The token embedding that represents an integer during pre-training will be used to represent a different and unrelated token in English.

\begin{table*}[h!]
\centering
\begin{tabular}{cccccccccc}
\hline
\textbf{L1} & \textbf{STS-B} & \textbf{QNLI} & \textbf{QQP} & \textbf{SST-2} & 
\textbf{MNLI} & \textbf{MRPC} & \textbf{Avg}\\
\hline
\textbf{Scratch}  & 18.6 (1.4) &62.1 (0.5) & 77.6 (0.2) & 82.2 (0.5)  & 62.1 (0.5) & 70.6 (3.5) & 61.6\\
\hline
\hline
\textbf{Pre. En} & +65.1 (1.0) & +22.4 (1.0) & +7.0(0.2) & +4.3 (0.3) & +12.7 (0.5) & +11.3 (0.5) & +20.5\\
\textbf{Pre. Ka} & +55.0 (1.5) & +14.7 (0.5) & +4.2 (0.1) & +0.3 (0.7) & +5.6 (0.0) & +8.5 (0.4) & +14.7\\
\hline
\textbf{Uniform} & -1.3 (0.1) & -1.7 (0.6) & -0.1 (0.2) & -0.3 (0.4) & +0.9 (0.2) & +1.4 (0.8) & -0.2\\
\textbf{Uni-gram} & +2.9 (1.6) & -0.6 (0.5) & +0.5 (0.2) & -1.0 (0.3) & +0.9 (0.1) & +5.1 (1.1) & +1.3\\
\textbf{Bi-gram} & +5.0 (1.7) & +0.0(0.3) & -0.4 (0.1) & -0.5 (0.4) & +1.4 (0.3) & +7.7 (0.9) & +2.2 \\
\hline
\textbf{Flat-2} & +19.9 (6.0) & +13.7 (0.1) & +0.8 (0.1) & -2.7 (0.1) & +3.2 (0.1) & +8.0 (1.4) & +7.2\\
\textbf{Flat-4} & +49.4(4.2) & +16.2 (0.4) & +1.5 (0.1) & -0.9 (1.4) & +4.7 (0.4) & -6.6 (1.5) & +10.7\\
\textbf{Flat-6} & +55.3(0.0) & +16.3 (0.3) & +2.3 (0.2) & +0.3 (0.9) & +5.9 (0.3) & +1.0 (2.2) & +13.5\\
\textbf{Flat-128} & +58.5 (0.8) & +16.0 (0.4) & +2.2 (0.0) & -1.1 (0.2) & +4.1 (0.3) & +7.5 (0.8) & +14.5\\
\textbf{Nest Par.} & +43.4 (4.3) & +17.3 (0.2) & +3.3 (0.3) & -1.1 (0.7) & +6.3 (0.3) & +3.2 (0.9) & +12.0\\
\hline
\textbf{Shuff.-64} & +49.7 (0.3) & +13.5 (0.3) & +1.5 (0.2) & -1.7 (0.7) & +3.3 (0.5) & +8.4 (0.3) & +12.4 \\
\textbf{Shuff.-32} & +44.7 (0.2) & +14.2 (0.7) & +1.3 (0.2) & -0.9 (0.5) & +3.4 (0.1) & +8.5 (1.3) & +11.9\\
\textbf{Shuff.-16} & +34.0 (0.2) & +14.6 (0.2) & +2.1 (0.1) & -3.1 (0.4) & +3.7 (0.4) & +7.2 (1.4) & +9.8 \\
\textbf{Shuff.-8} & +19.8 (7.8) & +13.0 (0.2) & +2.5 (0.2) & -0.0 (0.6) & +4.0 (0.2) & +5.8 (0.9) & +7.5 \\
\textbf{Shuff.-6} & +10.3 (0.4) & +13.8 (0.0) & +1.9 (0.1) & -1.2 (0.2) & +4.3 (0.1) & +9.3 (1.3) & +6.4 \\
\textbf{Shuff.-4} & +8.5 (2.3) & +9.0 (1.6) & +1.5 (0.1) & -0.4 (0.1) & +2.3 (1.0) & +7.1 (0.2) & +4.7\\

\hline
\end{tabular}
\caption{Downstream results of models in Section~\ref{subsec:baseline} to Section~\ref{sec:long-term}. 
We report the performance training from scratch on GLUE in the first block, and we report the relative improvement over the trained from scratch models for all other artificial datasets.
The value in the parentheses is the standard deviation.
The evaluation metrics of MRPC and QQP are F1 score, Spearman correlation coefficient is reported for STS-B, and the rest tasks are evaluated with accuracy.
Pre. is the abbreviation of pre-training, En stands for English, Ka stands for Kannada, Par. is short for Parentheses, and Shuff. is short for Shuffle.
}
\label{tab:relative}
\end{table*}

\section{Baseline Models}
\label{subsec:baseline}
In this paper, we use three different baseline performances to benchmark the performances pre-trained on different L1s.
\paragraph{English:} 
We pre-train a RoBERTa-medium using a subset of English Wikipedia.
The GLUE score obtained by fine-tuning this model serves as the performance upper bound for other models in our paper.
\paragraph{Kannada:}
We pre-train a RoBERTa-medium using Kannada from OSCAR dataset ~\citep{suarez2020monolingual}.
Kannada is a language spoken by the people in the southwestern region of India.
The main reason we choose this dataset lies in its subject(S)-object(O)-verb(V) structure, different from the S-V-O structure of our target language used in fine-tuning.
This model helps us understand how English downstream performance can benefit from pre-training on a non-English human language.

\paragraph{Training from scratch:}
We train transformer models with the same architecture of RoBERTa-medium directly on downstream GLUE tasks without pre-training.
This baseline performance, compared with other pre-trained models, helps us understand how effective pre-training is.

We show the performance of baseline models in the first and second block of Table~\ref{tab:relative}. 
Without any surprise, the model pre-trained on English performs the best, and the model trained from scratch on GLUE performs the worst, while the performance of the model transferred from Kannada is in between the other two baseline models.

\section{Characteristic 1: Matching the Distribution between Pre-training and Fine-tuning}
\label{sec:distribution}
The first characteristic we examine is the token distribution of the pre-training data.  
We construct three artificial datasets with different token distributions to understand how the token distributions for pre-training affect downstream GLUE performance after fine-tuning.

\subsection{Datasets}

\paragraph{Uniform}
The first artificial dataset is constructed by sampling 29995 integers from the vocabulary based on the uniform distribution to form a sequence.
The model learns nothing but randomly picks a token during pre-training.

\paragraph{Uni-gram}
The second dataset, called \textit{Uni-gram}, is designed to match the uni-gram token distribution of the real English token distribution.
This dataset is constructed by calculating the token distribution of the English Wikipedia dataset we used as the baseline, follows by sampling tokens based on that distribution to form sequences of variable lengths. 
Being able to perform MLM pre-task over this dataset indicates the model learns to model the downstream tasks' token distribution.

\paragraph{Bi-gram}
The third dataset is constructed such that the bi-gram distribution of this dataset matches that of the English Wikipedia corpus.
This dataset is constructed by sampling tokens based on the bi-gram distribution of the English Wikipedia subset.

\subsection{Results}
The results are in the third block of Table~\ref{tab:relative}.
We see that the model pre-trained with uniform token distribution performs the worst, which is as bad as the models trained from scratch.
The other two models that have learned the downstream tokens' distribution perform marginally better, while still falling far behind the two baseline models trained on human languages.
This implies that only modeling the downstream task's uni-gram distribution or bi-gram distribution is not enough to make a pre-trained language model able to perform well on the downstream tasks.

\section{Characteristic 2: Explicit Token Dependencies in a Sequence}
\label{sec:explicit}
In this section, we intend to focus on the \textit{explicit dependencies} between tokens in the sequences of the pre-training data.
In this paper, we say there exists an \textit{explicit dependency} between two tokens \(x_i\) and \(x_j\) if knowing one of the two tokens can tell us what the other token is.
Explicit token dependency is rich in human languages, such as the subject-verb agreement: if the verb is in singular form, then the subject must be singular.
We ask whether the extraordinary downstream performance of a pre-trained LM springs from its skillfulness in modeling the explicit dependencies among tokens in a sequence.
To this end, we construct datasets with explicit token dependencies.

\subsection{Data}
\begin{figure}[t]

\begin{subfigure}{0.5\textwidth}
\begin{center}
\includegraphics[width=0.8\textwidth]{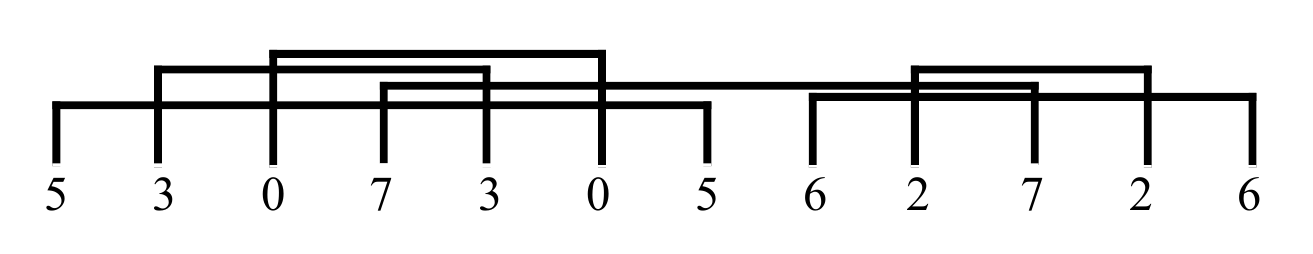}
\caption{Flat Parentheses (Section~\ref{sec:explicit})}
\label{fig:flat}
\end{center}
\end{subfigure}

\begin{subfigure}{0.5\textwidth}
\begin{center}
\includegraphics[width=0.8\textwidth]{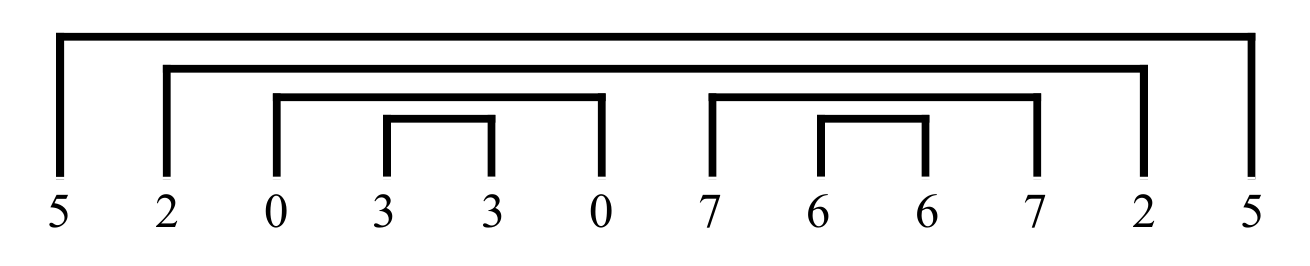}
\caption{Nesting Parentheses (Section~\ref{sec:explicit})}
\label{fig:nesting}
\end{center}
\end{subfigure}

\begin{subfigure}{0.5\textwidth}
\begin{center}
\includegraphics[width=0.8\textwidth]{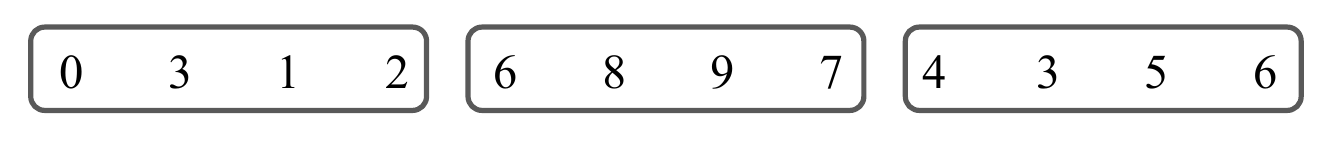}
\caption{Shuffle-4 (Section~\ref{sec:long-term})}
\label{fig:local4}
\end{center}
\end{subfigure}

\caption{Illustration of the artificial datasets in Section~\ref{sec:explicit} and~\ref{sec:long-term}. 
The arcs in~\ref{fig:flat},~\ref{fig:nesting} and blocks in~\ref{fig:local4} are showed here for easier understanding.}
\label{fig:data}
\end{figure}

\paragraph{Flat Parentheses}
The first dataset that contains explicit token dependencies is called \textit{Flat Parentheses}, and the reason for its name will be clear later.
We construct this dataset by first determining a half sequence length \(T/2\), and sample \(T/2\) (not necessarily consecutive) integers from English's uni-gram token distribution to form a sequence.  
We duplicate each token in the sampled sequence and then shuffle the sequence to form our final data with length \(T\).
Visualization can be found in Figure~\ref{fig:flat}.
Each integer in the generated sequence will occur even times, so we can view the same integers in the sequence as a pair of \textit{depending integers}.
We call the distance between a pair of depending integers in the sequence the \textit{length of dependency}.

To understand how the length of dependency in the pre-training data affects the fine-tuned downstream performance, we construct different \textit{Flat Parentheses} datasets that have a different maximum length of dependency.
We use \textit{Flat Parentheses-\(L\)} to denotes the \textit{Flat Parentheses} dataset that has a maximum dependency length of \(L\). 

\paragraph{Nesting Parentheses}
In the \textit{Flat Parentheses} dataset, if we connect each pair of depending integers with an arc as in Figure~\ref{fig:flat}, we can see that these arcs might cross with each other.
A special case of \textit{Flat Parentheses} is when all the arcs connecting all pairs of depending integers do not cross with each other, forming a hierarchical (nesting) structure, as in Figure~\ref{fig:nesting}.
This dataset is called \textit{Nesting Parentheses} in~\citet{papadimitriou-jurafsky-2020-learning}. 
We follow~\citet{papadimitriou2020learning} to generate this dataset by a stack-based grammar; the detailed procedure is in the Appendix~\ref{app:nesting}.
Figure~\ref{fig:nesting} shows a simple example.
We can observe from Figure~\ref{fig:nesting} that a sequence generated in this manner contains a nesting hierarchical parentheses structure, which is similar to the dependency tree structure in natural language.

\subsection{Results}
The results for the \textit{Parentheses} datasets are presented in the forth block of Table~\ref{tab:relative}.
Comparing with the results in the first and third block of Table~\ref{tab:relative}, it is clear that the \textit{Parentheses} datasets result in far better average GLUE scores than training from scratch on downstream tasks, or transferring from randomly generated sequences in Section~\ref{sec:distribution}.
It is interesting to note that the \textit{Flat Parentheses} dataset is quite similar to the \textit{Uni-gram} dataset in Section~\ref{sec:distribution} since tokens in both datasets are sampled from the same uni-gram distribution. The only difference is that the occurrence of each token in \textit{Flat Parentheses} must be even.
While bearing significant similarity, their performance diverges largely.
This indicates that learning to model the explicit token dependencies between tokens is critical to how well the pre-trained model can perform on downstream tasks.
Another astonishing observation is that pre-training on both \textit{Parentheses} datasets yields GLUE scores comparable to the model trained on another human language, Kannada.

The results in the fourth block show that the RoBERTa models trained on both ~\textit{Nesting Parentheses} and \textit{Flat Parentheses} have similar performances on downstream tasks. 
While we expect the model pre-trained on \textit{Nesting Parentheses} to learn to model hierarchical structure among tokens, the downstream performance does not surpass the model trained on \textit{Flat Parentheses}, which does not learn the hierarchical structure in pre-training.
The result implies that being skilled at modeling the hierarchical structure among tokens might not be critical to a pre-trained LM's transferability.

Comparing the results of \textit{Flat Parentheses-N}, we observe that when \(N=2\), their performance already outperforms that of the baselines trained on \textit{Uni-gram} or \textit{Bi-gram}.
While \textit{Flat Parentheses-2} degenerates to only repeating each token twice, pre-training on this trivial dataset still makes the pre-trained model performs better than other baseline models.
While \textit{Flat Parentheses-2} already outperforms models pre-trained on  \textit{Uni-gram} and \textit{Bi-gram}, its performance still falls behind other \textit{Flat Parentheses} datasets with longer dependency length.
This indicates that the knowledge transferred from datasets with explicit token dependencies is beneficial to English downstream tasks, and the longer dependency length the pre-training data has, the better downstream performance can be acquired.

\section{Characteristic 3: Implicit Token Dependencies in a Sequence}
\label{sec:long-term}
In this section, we focus on the \textit{implicit dependencies} among tokens presented in pre-training data.
In natural languages, a token in a sequence may implicitly depend on multiple neighboring tokens in the sequence.
For example, consider the following sentence: \textit{I can't believe I spent two hours watching that [MASK] movie.}
We might expect the \textit{[MASK]} to be a negative sentimental word, and we make this inference based on all the tokens in the sentence.
The dependency that a token depends on a set of neighboring tokens instead of a specific token is called \textit{implicit dependency} in this paper.
In terms of downstream performance, we want to know how important it is for a pre-trained LM to learn the implicit token dependencies among tokens in the sequence.
We are also interested in how the length of this implicit dependency affects downstream performance. 
We design a dataset called \textit{Shuffle-N} to answer the previous questions.

\subsection{Shuffle-N}
We explain how to construct this dataset, while the readers can refer to the example of \textit{Shuffle-4} in Figure~\ref{fig:local4}.
Each sequence in this dataset is formed by concatenating blocks of \(N\) integers.
In each block, we sample \(N\) consecutive numbers from 0 to 29994, and we shuffle the sampled integers to form a block. 
Integers in different blocks are sampled independently.
To solve the MLM task on this dataset, the model only needs to focus on a context that is no more than \(2N-1\) tokens (the previous \(N-1\) tokens and the next \(N-1\) tokens).
The maximum \(N\) we used in our experiments is \(64\), since the maximum sequence length in our model is \(128\).
By varying \(N\), we can examine how the length of the implicit dependency in the pre-training data affects the downstream performance.

Note that we do not require that each block to have non-overlapping integers when constructing the dataset.
Chances are that the same integer may occur in the same sequence multiple times, forming an explicit dependency.
However, we find that the generated datasets contain few explicit token dependencies, so learning this explicit token dependency does not help the model perform MLM.
We also carry out further analysis on Section~\ref{sec:analysis} to show that the models learned from this dataset indeed focus on the \(2N\) neighborhoods.

This dataset is designed to capture the constituent structure of human language, and different N can capture the variable length that a constituent may span in human language. 
Since words in the same constituent have some implicit dependency, we hypothesize that LMs trained on human language will learn how to model the dependency among the tokens in it, leading to the superior transferability of natural language trained LMs.
In Shuffle-N, the ‘block’ corresponds to the ‘constituent’ in human language, and those tokens in the block correspond to words in a constituent. 

\subsection{Results}
The results for \textit{Shuffle-N} is presented in the fifth block in Table~\ref{tab:relative}.
We find that all models pre-trained on \textit{Shuffle-N} yield better downstream performance than the \textit{Uni-gram} and \textit{Bi-gram}, showing that learning to model implicit dependency among tokens during pre-training is critical for the model to perform well on downstream tasks.
Comparing the performance of \textit{Shuffle} with different \(N\), we observe that averaged performance increases as \(N\) increases.
Most GLUE tasks show improvement when \(N\) increases from \(4\) to \(6\), while STS-B's performance consistently goes up as \(N\) increases to \(64\).
While we yet to know the reason behind the positive correlation between the downstream STS-B's performance and the pre-train implicit dependency length \(N\), it is still surprising to see that only varying a specific characteristic of the pre-training data can make such a great difference on a downstream task's performance.

\begin{figure}[h!]
\centering
\includegraphics[width=0.85\linewidth]{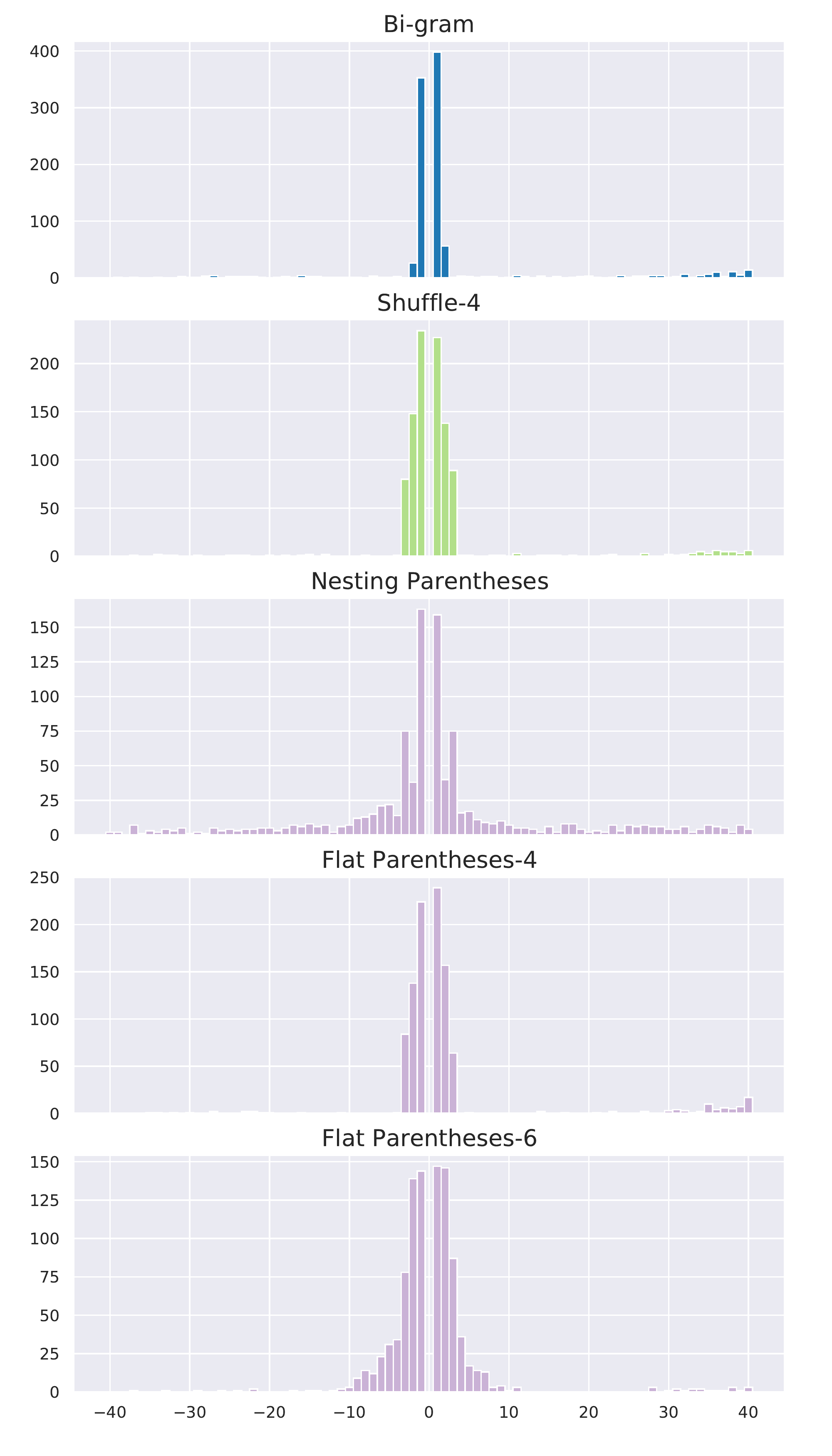}
\caption{The distribution of \(j^*\). The x-axis is the relative position with respect to the \(\lfloor{T/2}\rfloor\)-th token in the input sequence.
The height of the histogram when x-axis value is \(j\) represents the number of count that \(j=j^*\) among sentences in SQuAD.}
\label{fig:attn}
\end{figure}

\section{Analysis: How Does Different Pre-training Artificial Datasets Affect the Models' Behavior on Human Language?}
\label{sec:analysis}

In the previous three sections, we use different artificial datasets to understand how a specific trait in the pre-training data affects the downstream performance.
While we anticipate that the model pre-trained with an artificial dataset with a certain characteristic will learn specific ways to process the artificial input data during pre-training, it is unclear whether the model will have a similar behavior when processing a human language.
In this section, we design an experiment to analyze how the models pre-trained on artificial datasets behave when faced with human language.

\subsection{Method}
During MLM pre-training, the model needs to reconstruct a masked token based on the remaining tokens in the sequence.
Given an input sequence \(X\) with \(T\) tokens, let \(x_i\) denotes the i-th token in \(X\), and let \(X_{\backslash\{i,j\}}\) denotes the sequence \(X\) with \(x_i\) and \(x_j\) being masked.
We would like to know how the presence of \(x_i\) affects how well the model can predict \(x_j\).
To that end, we compare the LM's prediction at the j-th position when the input sequence is \(X_{\backslash\{j\}}\) and \(X_{\backslash\{i,j\}}\).
Since the LM's prediction is a probability distribution over the set of all tokens in the vocabulary, we can calculate the entropy of the model's prediction.
We use \(P_{\backslash\{j\}}\) and \(P_{\backslash\{i,j\}}\) to denote the LM's prediction at the j-th position when the input sequence is \(X_{\backslash\{j\}}\) and \(X_{\backslash\{i,j\}}\), and denote the entropy of a distribution \(P\) as \(H(P)\).
If knowing \(x_i\) is important for the prediction of \(x_j\), then \(H(P_{\backslash\{j\}})\) should be lower than \(H(P_{\backslash\{i,j\}})\) since knowing \(x_i\) will reduce the uncertainty for predicting \(x_j\).

In our experiment, given a sequence \(X\), we choose the \(\lfloor{T/2}\rfloor\)-th token in the input sequence as \(x_i\) and iterate over all the tokens in the sequence as \(x_j\) to find which position \(j\) is most related to the \(\lfloor{T/2}\rfloor\)-th token as below.
\begin{equation}
j^*=\arg\max_{j:x_j\in X} H(P_{\backslash\{\lfloor{T/2}\rfloor,j\}})-H(P_{\backslash\{j\}}).
\end{equation}
Then we find the \(j^*\) for each input sentence and accumulate the distribution of \(j^*\) for all sentences in a dataset.

\subsection{Result}
The sentences used for computing the distribution of \(j^*\) here are from SQuAD~\citep{rajpurkar2016squad}\footnote{We choose SQuAD instead of GLUE since sentences in SQuAD are longer.}.
The results are in Figure~\ref{fig:attn}.
First, we immediately find that the \(j^*\) distribution of \textit{Bi-gram} peaks around the center. 
This is expected since during pre-training, the model trained on \textit{Bi-gram} learns to only focus on a very small context.
We can observe that both the distributions of \textit{Flat Parentheses-6} and \textit{Flat Parentheses-4} center near the \(\lfloor{T/2}\rfloor\)-th token, which is the expected behavior.
Furthermore, the distribution of \textit{Flat Parentheses-4} is more condensed around the \(\lfloor{T/2}\rfloor\)-th token than that of \textit{Flat Parentheses-6}, showing that pre-training on the former dataset makes the model focus on a smaller context.
Another interesting observation is that the distribution of \(j^*\) for \textit{Nesting Parentheses} shows a high-low-high-low pattern around the \(\lfloor{T/2}\rfloor\)-th token, this phenomenon springs from the nature of the \textit{Nesting Parentheses} dataset.
The distribution of \(j^*\) for \textit{Shuffle 4} is also focused within the previous 3 tokens and the next 3 tokens of the \(\lfloor{T/2}\rfloor\)-th token, which also results from the nature of the dataset the model is pre-trained on.
We thus verify that what we expect the model to learn from the artificial dataset can be transferred to English downstream tasks. 

\begin{table*}[ht!]
\centering
\begin{tabular}{ccccccc}
\hline
\textbf{Tasks} & \textbf{Pre-train En} & \textbf{From Scratch} & \textbf{Pre-train Ka}  & \textbf{Bi-gram}& \textbf{Nest.Par.} & \textbf{Shuffle-4}\\
\hline
\textbf{QQP} & 84.6 & 77.6 & 81.8 & 77.2 & 80.9 & 79.1\\
\textbf{QQP\textsubscript{PAWS}} & 43.0 & 43.9 &44.0 & 42.7 & 43.0 & 42.8\\
\hline
\textbf{QNLI} & 84.5 & 62.1 & 76.8 & 59.8 & 79.4 & 71.1\\
\textbf{QNLI-Adv} & 64.8 & 0.1 & 8.6 & 53.8 & 1.0 & 1.1\\
\hline
\end{tabular}
\caption{Evaluation performance of models fine-tuned on QQP and QNLI.
The evaluation metrics are the same as Table~\ref{tab:relative}.
}
\label{tab:adversarial}
\end{table*}
\section{How Robust are Models Pre-trained with Different Datasets?}
\label{sec:adversarial}
We have evaluated the pre-trained models trained with artificial datasets of different traits, and the results show that some models are as good as the models pre-trained from a human language in terms of GLUE scores. 
Are these models as robust as the models pre-trained from human language? 
Chances are the GLUE scores on the evaluation set are decent, but the model only learns to fit some spurious correlation of the GLUE tasks, making the model less robust.
Here we use two challenging datasets to assess the robustness of the fine-tuned models pre-trained on different datasets.

\subsection{Experiment Setup}
We use RoBERTa models pre-trained on different L1s, fine-tuning them using the \textbf{original} GLUE training set, and test the fine-tuned model on a more challenging evaluation set.
We use two GLUE tasks, QQP and QNLI, and their corresponding challenging datasets, QQP\textsubscript{PAWS} and QNLI-Adv.
\paragraph{QQP\textsubscript{PAWS}}
The Quora Question Pairs (QQP)~\citep{iyer2017first} contains real-world question pairs collected from Quora. 
While this dataset is widely used to train and evaluate a paraphrase model, it is shown that paraphrase pairs in QQP tend to contain high lexical overlap.
This makes us hard to determine whether a model really learns how to distinguish between paraphrases and non-paraphrases, or it just learns the spurious correlation that similar bag-of-word implies paraphrase. 

QQP\textsubscript{PAWS} is proposed by ~\citet{zhang2019paws} to solve the above problem.
~\citet{zhang2019paws} use word swapping and back translation along with human annotation to create a high-quality paraphrase dataset.
In QQP\textsubscript{PAWS}, two questions with high lexical overlap do not often imply they are a pair of paraphrases.

\paragraph{QNLI-Adv}
In the original Question-answering NLI (QNLI), given a question-answer pair, the model needs to determine whether the answer corresponds to the question, i.e., entailment.
To see whether the model only learns to label question-answer pairs with high lexical overlap as entailment, we propose a challenging evaluation dataset from the original QNLI evaluation split.
We call this dataset~\textit{QNLI-Adv}.
The construction of \textit{QNLI-Adv} is extremely simple: instead of feeding the model a pair of question and answer, we give the model two identical questions, and the model has to answer non-entailment to this input.
That is, the ground-truth labels of data in QNLI-Adv are all non-entailment.

\subsection{Result}
From Table~\ref{tab:adversarial}, we observe that all pre-trained models are equally vulnerable to QQP\textsubscript{PAWS}. 
This may result from the fact that QQP\textsubscript{PAWS} is deliberately constructed to be a challenging dataset and has a low discrimination index.

On the other hand, when evaluated with QNLI-Adv, the models behave differently. 
The model pre-trained on English is the most robust one, with the accuracy only dropping 20\% comparing to the original QNLI evaluation set.
We find that both ~\textit{Nesting Parentheses} and ~\textit{Shuffle-4} performs unbelievably poorly, showing that while learning to model the explicit and implicit token dependencies enables the model to perform well on downstream tasks, it is vulnerable to spurious correlations in the downstream tasks. 
This vulnerability against spurious correlation can not only be observed on model pre-trained on artificial datasets; even the model pre-trained on Kannada, a human language, performs poorly on QNLI-Adv, indicating that not learning the semantics of the downstream language will make the model less robust toward challenging datasets.
The unreasonably high robustness of \textit{Bi-gram} may be attributed to the fact that the performance of \textit{Bi-gram} on the original evaluation set is not much better than random guessing among entailment and non-entailment, so randomly guessing will result in accuracy around 50\%.

\section{Discussion and Conclusion}
In this work, we study what traits \textit{besides semantics} in the pre-training data make the pre-trained LMs able to yield exceptional downstream performance.
We propose to study this problem with the aid of artificial datasets.
The framework is general, and thus if one would like to study whether other characteristics will affect the transferability of transformer-based LM, they can adopt this framework.

Specifically, we construct linguistic-inspired artificial datasets, and finding that pre-training on certain artificial datasets makes the MLMs' English downstream performance comparable to transferring from an MLM pre-trained on an non-English human language.
We show that both the explicit and implicit dependencies between tokens in the sequences are critical to the transferability of the pre-trained model.
The result in Tabel~\ref{tab:relative} indicates that pre-training on artificial datasets with explicit/implicit token dependencies makes the pre-trained LMs superior to the from-scratch and Uni/Bi-gram baselines. 

The downstream performance of LMs pre-trained on datasets with explicit/implicit token dependencies still falls behind models pre-trained with English.
This performance gap is expected: when using artificial datasets to pre-train an LM, the LM cannot learn any semantic features useful for downstream tasks.
We also carry out experiments to test the models' behavior when faced with challenging datasets and showing that models pre-trained without English are more prone to learn spurious correlation of the downstream tasks.

Overall, our results contribute to an important problem in the NLP community: where does the transferability of pre-trained transformers arise from? 
While one may infer that transformers pre-trained on natural language can model token dependencies in the sequences, it is unclear how much this contributes to the transferability of the pre-trained transformer LMs.
We disentangle the effect of semantic similarity during pre-training and downstream fine-tuning. 
We show that even when the pre-training data and downstream tasks share no semantic features, the transformer LMs possess positive transferability to natural language downstream tasks if it has the ability to model the token dependencies in the sequences.
We attribute the transferability of pre-trained transformer LMs to their capability of modeling the dependencies among tokens, and we envision that the results may help researchers in different disciplines to apply transformer pre-trained models to their domains of interests.


\section*{Acknowledgement}
We want to thank Wei-Tsung Kao, Yist Y. Lin, and Yung-Sung Chuang for their valuable feedbacks on the draft of our paper.
We also want to thank the anonymous reviewers for providing insightful and actionable suggestions on our work.
We thank National Center for High-performance Computing (NCHC) of National Applied Research Laboratories (NARLabs) in Taiwan for providing computational and storage resources.

\bibliography{aaai22}

\clearpage
\appendix
\section{Experiment Details}
\label{app: pretrain}
We give detailed model architectures of our RoBERTa-medium model and hyperparameters used in pre-training.
\subsection{Model}
We use RoBERTa-medium, a 8-layered transformer model with hidden dimension 512 and 8 attention heads per layer. 
The total number of parameters of the model is around 41M.
We pre-train RoBERTa using Huggingface~\citep{Wolf2019HuggingFacesTS} code base.
We do not perform early stopping during pre-training or fine-tuning.
Pre-training and fine-tuning a model takes 20 hours on a single V100.
\subsection{Hyperparameters}
The hyperparameters used in all pre-training experiments are listed in Table~\ref{tab:pretrain hyperparam}
\begin{table}[h]
    \centering
    \begin{tabular}{|c|c|}
    \hline
        Batch size & 300 \\
        Learning rate & 5E-5 \\
        Total steps & 100K \\
        Warmup steps & 5k\\
        Max Position & 128\\
    \hline
    \end{tabular}
    \caption{Pre-training hyperparemeters for BERT.}
    \label{tab:pretrain hyperparam}
\end{table}
\subsection{Pre-training Data}
We put all details related to all pre-training data in Table~\ref{tab:pretrain data}.
We provide download link to the pre-training dataset.
The artificial data can be generated following the script in our code.
We also include the vocabulary size (including special tokens) of each model on the last column.
The vocabulary file is obtained by training a WordPiece tokenizer on the training data for Kannada, and Wikipedia dataset.

\begin{table}[h]
    \centering
    \begin{tabular}{ccc}
    \hline
        \textbf{Dataset} & \textbf{Number of Tokens} & \textbf{Vocab Size}\\
        \hline
        Wikipedia & 118M&30000\\
        Kannada  &77.5M&30000\\
        \hline
        All artificial datasets & 100M & 30000\\
    \hline
    \end{tabular}
    \caption{Details for dataset used in pre-training.}
    \label{tab:pretrain data}
\end{table}
\subsection{Fine-tuning Details}
We fine-tune GLUE using Huggingface~\citep{Wolf2019HuggingFacesTS} code base.
The model fine-tuned in this section is RoBERTa-medium with classifier on top of the last transformer layer.
The whole model fine-tuned is has 41M parameters.
\subsubsection{Dataset}
We provide statistics on the 8 GLUE tasks we used in Table~\ref{tab:downstream_st}
\begin{table}[]
    \centering
    \begin{tabular}{c|c}
       Task  & Examples \\
       \hline
       MRPC & 3.6K / 0.4K / 1.7K \\
       STS-B & 5.7K / 1.5K / 1.3K \\
       QNLI & 104K / 5.4K / 5.4K \\
       QQP & 363K / 40.4K / 391.0K\\
       MNLI & 392.7K / 9.8K + 9.8K / 9.8K + 9.8K\\
       SST-2 & 67.4K / 0.9K / 1.8K\\
    \end{tabular}
    \caption{Statistics of (train / dev/ test) in GLUE tasks
    MNLI contains matched and mismatched in dev and test set. We didn't evaluate our models' performance on test set.}
    \label{tab:downstream_st}
\end{table}
\subsubsection{Fine-tuning Hyperparameters}
We list the hyperparameters used in fine-tuning GLUE in Tabel~\ref{tab:glue_hpp}.
\begin{table*}[ht]
    \centering
    \begin{tabular}{c|cccccccc}
    & LR & BSZ & RoBERTa DR & Classifier DR & TS & WS & MSL \\
    \hline
    STS-B &2.00E-05& 16 &0 &0.1 &3598& 214 &128\\
    SST-2& 1.00E-05 &32& 0& 0.1 &20935 &1256 &128 \\
MNLI& 3.00E-05 &128& 0 &0.1 &10000 &1000& 128\\
QNLI& 1.00E-05 &32& 0 &0.1& 33112& 1986& 128\\
QQP& 5.00E-05& 128& 0 &0.1& 14000 &1000 &128\\
MRPC& 2.00E-05& 32& 0 &0.1& 800& 200 &128\\
    \end{tabular}
    \caption{Hyperparameters for RoBERTa in downstream tasks. LR: Learning Rate. BSZ: Batch Size. DR: Dropout Rate. TS: Training Steps. WS: Warmup Steps. MSL: Maximum Sequence
Length}
    \label{tab:glue_hpp}
\end{table*}
\section{Nesting Parentheses}
\label{app:nesting}
This dataset is generated by the following stack-based grammar, following ~\citet{papadimitriou2020learning}: At each time step \emph{t}, we sample \(X_t\) from a Bernoulli distribution with \(P(X_t = 1) = 0.4\).
If \(X_t=1\), we sample a token based on English's uni-gram distribution, place the sampled token at position \(t\) of the generated sequence, and push the same token into the stack.
When \(X_t=0\), we pop the top element of the stack and put the popped token at position \(t\) of the generated sequence.
\section{Negative Results}
We also tried to initialize the word embedding layer of RoBERTa while keeping all other parameters when fine-tuning on GLUE.
That is, the transformer layers of RoBERTa inherit the parameters from pre-training on artificial datasets, and we randomly initialize the word embedding layer; then the whole model is fine-tuned on downstream GLUE tasks.
However, we find that randomly initializing parameters make the model loss is ability to be fine-tuned.
The model performs as worse as models trained from scratch on downstream tasks.

\section{FAQs}
\begin{itemize}
    \item \textbf{Q1 } How general are the results in the paper with respect to the model size?
    \item \textbf{A1 } We’ve run experiments using the RoBERTa-base models (12-layered) in our preliminary experiments, and find that all our observation in this paper holds, while performance is slightly better compared with RoBERTa-medium. We choose medium model because it’s smaller and faster.
    We conclude that our results are general with respect to the size of the model.
    \item \textbf{Q2} How general are the results in the paper with respect to the different transformer architectures or pre-trained models?
    \item \textbf{A2} We only use encoder-based transformer models (RoBERTa, which has the exactly same architecture as BERT and the Electra’s discriminator), and we did not use the encoder-decoder models (e.g., BART or T5). 
    Given that the auxiliary tasks (next sentence prediction in BERT, sentence order prediction in ALBERT, and none in RoBERTa) do not significantly alter the pre-training results on natural languages, we also believe that our results will be robust and general to the selection of MLM pre-training objectives.
    In our future work, we plan to extend the observation in this paper to BART-like model.
    \item \textbf{Q3 } How do the authors assign the integers during pre-training to the sub-word tokens in fine-tuning?
    \item \textbf{A3 } We use two different methods to assign downstream token ids from pre-training. 
    The first one is random assignment, and the second is assigning the token ids based on the tokens frequency. 
    For example, during pre-training, token id 10 corresponds represents the token frequency of the 10-th token in English uni-gram distribution; then during fine-tuning, token id 10 will be used to represent the 10-th token in English uni-gram distribution.
    We find that no matter we use different random assignments or assignment based on token frequency, our observation does not vary. 
    We did not try to assign the token ids to the vocabulary based on how they appear together during pre-training. 
    This is mainly because that, in the case of implicit dependencies, consecutive ids will occur locally (which is deterministic); however, during fine-tuning, tokens in English do not always co-occur based on certain rules. 
    We thus are not able to come up with methods to map the pre-training token ids to the English tokens.
    \item \textbf{Q4 } Why QNLI-adv is constructed that way?
    \item \textbf{A4 } Our goal is to understand if the model performs QNLI based on the lexical overlap between question(Q)/answer(A). 
    (In the original QNLI, the goal is to determine whether A answers Q.) 
    We construct an extreme case: Making the Q and A have 100\% overlapping. 
    If the A’ is just the copy of Q, A’ does not answer Q and should not be considered ’Q has been answered.
    We use this extreme case to observe whether model trained on original QNLI dataset only learns to capture lexical overlap between the question and the answer.
    \item \textbf{Q5 } Will the results be affected if the datasets in Section~\ref{sec:explicit} and Section~\ref{sec:long-term} are not sampled based on the English Uni-gram distribution?
    \item \textbf{A5 } We constructed the \textit{Flat Parentheses} and \textit{Nesting Parentheses} datasets using the uniform distribution in our preliminary experiment, and found that results are not statistically different from those obtained by sampling from Uni-gram distribution. 
    Thus, the presence of information from the the English corpus might not be useful.
    \item \textbf{Q6 } Does sharing the token indexing during pre-training during fine-tuning make semantic information transfer between pre-training and fine-tuning?
    \item \textbf{A6 } While token indices during pre-training overlap with those used during fine-tuning, overlapping of token indices does not indicate overlapping of semantics.
    Let us consider semantics as the ”meaning of a token”. 
    From this aspect, the tokens used during pre-training and fine-tuning share no overlap of their meaning. 
    This is because during pre-training, the sampling process does not grant any meaning to the token indices. The only ”semantic” (meaning) each token may bear is its frequency, and the numerical relationship between the indices (e.g., the token 3 precedes the token 4, or the token 3, 4, 5, 6 are ”consecutive”). 
    This ”semantic” during pre-training is very different from the semantic in the downstream task, where each token’s meaning can be represented by the composition of the other tokens.
    Overall, even the pre-trained model has learned the ”semantics of artificial dataset” during pre-training, those semantics are not related with the semantics of the downstream tasks. 
    This is why we claim that our experiment setting can exclude the effect of semantics (of human language) during pre-training. 
\end{itemize}

\end{document}